\documentclass[letterpaper]{article} 
\usepackage{aaai23}  
\usepackage{times}  
\usepackage{helvet}  
\usepackage{courier}  
\usepackage[hyphens]{url}  
\usepackage{graphicx} 
\usepackage{natbib}  
\usepackage{caption} 
\usepackage{algorithm}
\usepackage{algorithmic}
\usepackage{multirow}
\usepackage[svgnames]{xcolor}         
\usepackage{soul}

\newcommand{\hlc}[2][yellow]{{%
    \colorlet{foo}{#1}%
    \sethlcolor{foo}\hl{#2}}%
}
\usepackage{newfloat}
\usepackage{listings}
\DeclareCaptionStyle{ruled}{labelfont=normalfont,labelsep=colon,strut=off} 
\floatstyle{ruled}
\newfloat{listing}{tb}{lst}{}
\floatname{listing}{Listing}
\usepackage{xcolor}
\usepackage{amsmath, amssymb, amsfonts}
\usepackage{booktabs}
\usepackage{rotating}
\usepackage{tikz, pgf}
\usetikzlibrary{arrows, arrows.meta, shapes, intersections, backgrounds, shapes.geometric, petri}

\usepackage{subfigure}
\usepackage{tabularx}
\usepackage{lipsum}
\usepackage{xspace}
\usepackage{booktabs}
\newcommand{\etal}{et al.\@\xspace}

\newcommand{\vspacesection}{\vspace{-0.2em}}

\newcommand{\vspacereduce}{\vspace{-1em}}

\title{Importance of Synthesizing High-quality Data for Text-to-SQL Parsing}
\author{
    Yiyun Zhao\textsuperscript{\rm 1} \footnote{Work done during internship at Amazon},
    Jiarong Jiang\textsuperscript{\rm 2},
    Yiqun Hu\textsuperscript{\rm 2},
    Wuwei Lan\textsuperscript{\rm 2},
    Henry Zhu\textsuperscript{\rm 2},
    Anuj Chauhan\textsuperscript{\rm 2},
    Alexander Li\textsuperscript{\rm 2},
    Lin Pan\textsuperscript{\rm 2},
    Jun Wang\textsuperscript{\rm 2},
    Chung-Wei Hang\textsuperscript{\rm 2},
    Sheng Zhang\textsuperscript{\rm 2},
    Marvin Dong\textsuperscript{\rm 2},
    Joe Lilien\textsuperscript{\rm 2}
    Patrick Ng\textsuperscript{\rm 2},
    Zhiguo Wang\textsuperscript{\rm 2},
    Vittorio Castelli\textsuperscript{\rm 2},
    Bing Xiang\textsuperscript{\rm 2}
}
\affiliations{
    \textsuperscript{\rm 1}University of Arizona\\
    yiyunzhao@email.arizona.edu\\

    \textsuperscript{\rm 2} AWS AI Labs\\
    \{jiarongj, yiqunhu, lanwuwei, patricng\}@amazon.com\\
}

\usepackage{bibentry}

\begin{document}

\maketitle

\begin{abstract}
Recently, there has been increasing interest in synthesizing data to improve downstream text-to-SQL tasks. In this paper, we first examined the existing synthesized datasets and discovered that state-of-the-art text-to-SQL algorithms did not further improve on popular benchmarks when trained with augmented synthetic data. We observed two shortcomings: illogical synthetic SQL queries from independent column sampling and arbitrary table joins. To address these issues, we propose a novel synthesis framework that incorporates key relationships from schema, imposes strong typing, and conducts schema-distance-weighted column sampling. We also adopt an intermediate representation (IR) for the SQL-to-text task to further improve the quality of the generated natural language questions. When existing powerful semantic parsers are pre-finetuned on our high-quality synthesized data, our experiments show that these models have significant accuracy boosts on popular benchmarks, including new state-of-the-art performance on Spider.

\end{abstract}

\section{Introduction}

Text-to-SQL parsing refers to the semantic parsing task that translates a natural language question (NLQ) to a corresponding SQL query.
In recent decades, many industries have adopted high-level digitalization in their workflow and possessed large-scale datasets---many of which are stored as relational databases. Extracting insights from these relation databases to further drive business decisions is an important task. But due to the complexity of these relational databases, query language experts are often needed to extract valuable insights. Thus a high-performing text-to-SQL system with a natural language interface would greatly lower the barrier for users to query their databases.

In order to obtain high-quality training data for the text-to-SQL parser, human annotators with SQL expertise are needed to construct NLQ-SQL parallel data, which are difficult and expensive to scale. Thus data scarcity is a well-known bottleneck in the text-to-SQL task \citep{yu-etal-2018-spider}.
To address the data scarcity issue, there is an increasing interest in leveraging synthetic data to improve downstream performance.
\citet{yu2021grappa} handcrafted high-quality rules to synthesize SQL and NLQ simultaneously, but these grammar rules need to be carefully designed through expensive manual work. 
To automate the synthesis procedure, recent works \cite{wang-etal-2021-learning-synthesize,wu-etal-2021-data,shi2021learning,zhong-etal-2020-grounded} utilize a two-stage approach that synthesizes SQL first and then composes NLQ with a SQL-to-text generator. 
Alternatively, \citet{yang2021hierarchical} proposed a reversed pipeline that uses an entity-to-question model to generate natural language queries and then a text-to-SQL parser to generate SQL queries. 

In this paper, we delve into the two-stage synthesizing method that first synthesizes SQL queries and then generates NLQs. 
We first experimented with two recent synthetic datasets \cite{wang-etal-2021-learning-synthesize} and \cite{wu-etal-2021-data} using the latest state-of-the-art text-to-SQL model PICARD \cite{Scholak2021:PICARD}.
We chose these two synthetic datasets because both are recent work that demonstrated efficacy with the popular high-performing RAT-SQL parser \cite{wang-etal-2020-rat} on the Spider benchmark \cite{yu-etal-2018-spider}. Surprisingly, our experimental results revealed that these two recent synthetic datasets show only negligible impact on downstream accuracy when trained on the PICARD model in a data augmentation fashion. Our manual inspection identifies three main sources of noise in these synthetic datasets: (1) illogical synthetic SQLs due to invalid grammars, (2) complex SQLs due to arbitrary multi-table joins, and (3) language gap between SQL and NLQ. 



\begin{figure*}[h]
\centering
\includegraphics[width=\textwidth]
{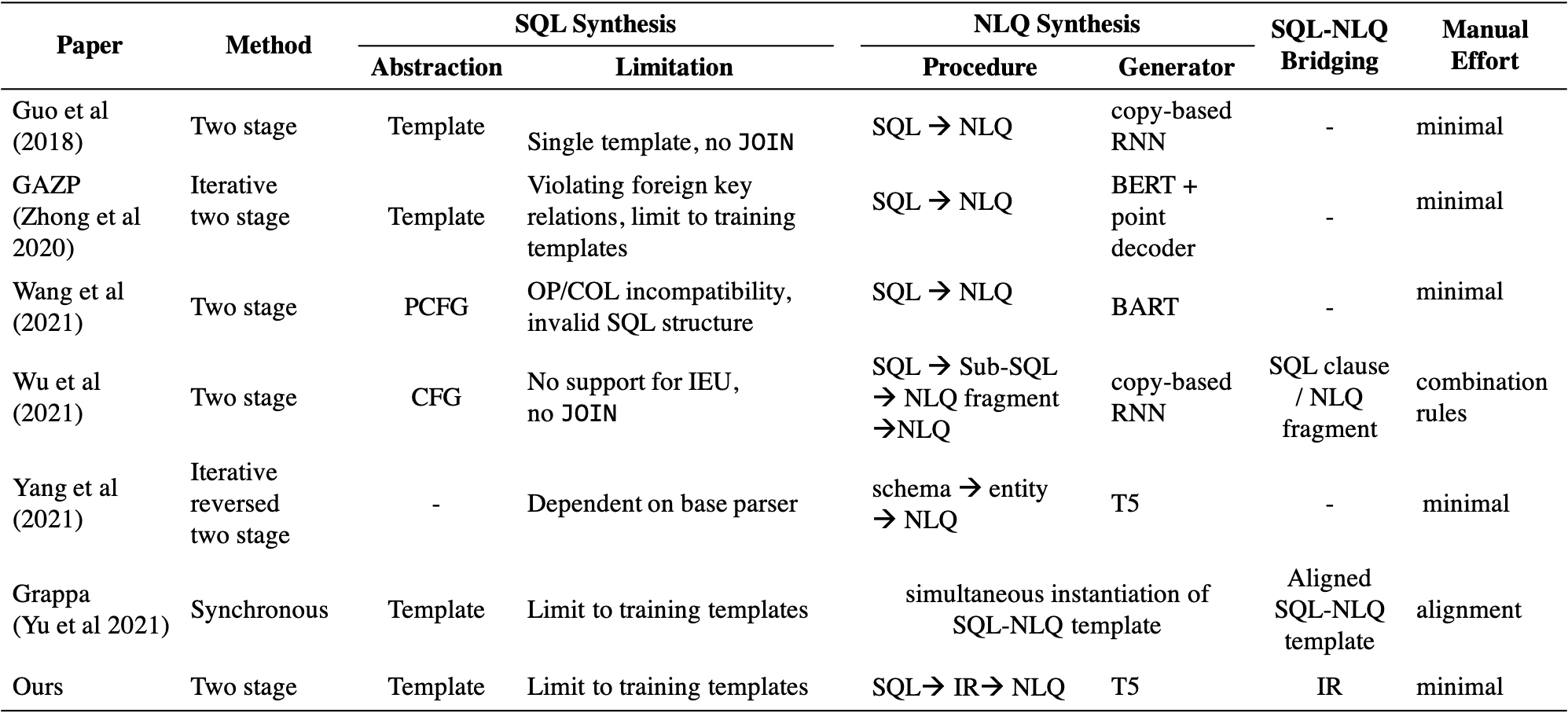}
\caption{Comparison of different data synthesis methods for text-to-SQL task.
\textit{Synchronous} refers to generating SQL and NLQ together, \textit{Two-stage} first synthesizes SQL then generates NLQ, \textit{reversed two-stage} first generates NLQ then synthesizes SQL. \textbf{SQL-NLQ Bridging} refers to intermediate operations or representations for matching SQL and NLQ.}
\label{fig:mini-survey}
\vspacereduce
\end{figure*}

We propose a novel framework\footnote{Source code will be made publicly available.} that has several strategies to reduce these synthesis errors. During the stage of SQL synthesis, we employ template synthesis with 
strong typing, template key relationship preservation, and schema-distance-weighted column sampling. 
During the stage of text generation, we propose an intermediate representation to bridge the gap between SQL queries and natural language questions. 
We show that models trained with our synthetic datasets outperform the models trained with previous synthetic datasets. Our model achieves new state-of-the-art accuracy on the Spider benchmark. 
In summary, our main contributions are:
\begin{itemize}
    \item We systematically compare the existing text-to-SQL synthesis methods and identify three root causes of low quality;
    \item We propose several novel strategies to improve data synthesis quality and demonstrate augmentation benefits when using the state-of-the-art PICARD parser, underscoring the importance of the synthesis quality; 
    \item We adopt an intermediate representation (IR) for the SQL-to-text task, which can further improve the quality of the generated natural language questions.
\end{itemize}



\subsection{Existing Synthesis Methods and Limitations}

Figure \ref{fig:mini-survey} presents the existing text-to-SQL synthesis methods and their characteristics from different dimensions. We detail each of them as follows:


Inspired by prior work by \citet{jia-liang-2016-data} in semantic parsing, \citet{yu2021grappa} extended a synchronous context-free grammar (SCFG) approach to the text-to-SQL task where they manually crafted about 90 high-quality SQL-NLQ aligned patterns to generate new SQL-NLQ pairs. They found pretraining on the synthetic dataset leads to a significant improvement even tested with a very strong text-to-SQL parser RAT-SQL on the Spider benchmark.

While SCFG usually creates high-quality data because patterns are carefully designed and aligned, the coverage of the patterns is limited, and expert knowledge is required to design such patterns. Thus, more efforts are devoted to automating the procedure. \citet{guo-etal-2018-question} utilized a two-stage approach by first sampling SQL queries from a simple pattern and then generating questions using a copy-based RNN encoder-decoder structure find the synthetic data that can improve the existing state-of-the-art model on the WikiSQL benchmark. \citet{zhong-etal-2020-grounded} followed the same two-stage approach but used templates extracted from training to generate SQL and augmented the NLQ generator with pretrained transformer BERT and iteratively updated the parser and generator. Only the synthetic dataset that was created using target schemas filtered with cycle consistency can facilitate the downstream performance. 

Along the same approach, \citet{wang-etal-2021-learning-synthesize} identified problems with fixed SQL synthesis rules and employed a full-fledged probabilistic context-free grammar (PCFG) that enabled generating SQLs with varying structures. They synthesized natural language queries with a BART SQL-NLQ generator. Their synthesis method has been shown to boost the RAT-SQL parser performance on the Spider benchmark, though the improvement is not as significant as pretraining using SCFG generated synthetic data \citep{yu2021grappa}. The gap might be due to the quality of the synthetic dataset as the independent selection of generation step in PCFG introduces substantial noise such as illogical SQL queries. 

To improve the quality of synthetic data, \citet{wu-etal-2021-data} introduced a clause-level synthesis framework: first decomposing a query into sub-clauses and translating sub-SQL clauses into sub-questions, and finally assembling sub-questions into a whole question. They found clause-based synthesis method is better than flat synthesis. 

Alternatively, \citet{yang2021hierarchical} proposed to improve the quality of synthetic data by incorporating domain information in question generation. Specifically, they learned an entity sampler and synthesized questions using an entity-to-question generator with entities sampled from the sampler, followed by generating pairing SQL queries through a baseline parser. For this approach, they also attractively updated the parser and generator, in a similar fashion as in \citet{zhong-etal-2020-grounded}. Their synthetic dataset can significantly improve a DT-Fixup parser on the Spider benchmark. 

This work seeks to investigate value of synthetic dataset with current state-of-the-art PICARD model and refine a synthetic method in an automate and non-iterative manner. Thus, we examine two synthetic datasets from recent work \cite{wang-etal-2021-learning-synthesize, wu-etal-2021-data} that demonstrate improvement of downstream performance with previous state-of-the-art text-to-SQL parser (RAT-SQL) over Spider benchmark without iterative training.

\subsubsection{Synthetic Data Effectiveness Assessment}
\label{session:assessmention_previous}
\begin{figure}[!h]
    \vspace{-1em}
     \centering
     \includegraphics[height=0.25\textwidth]{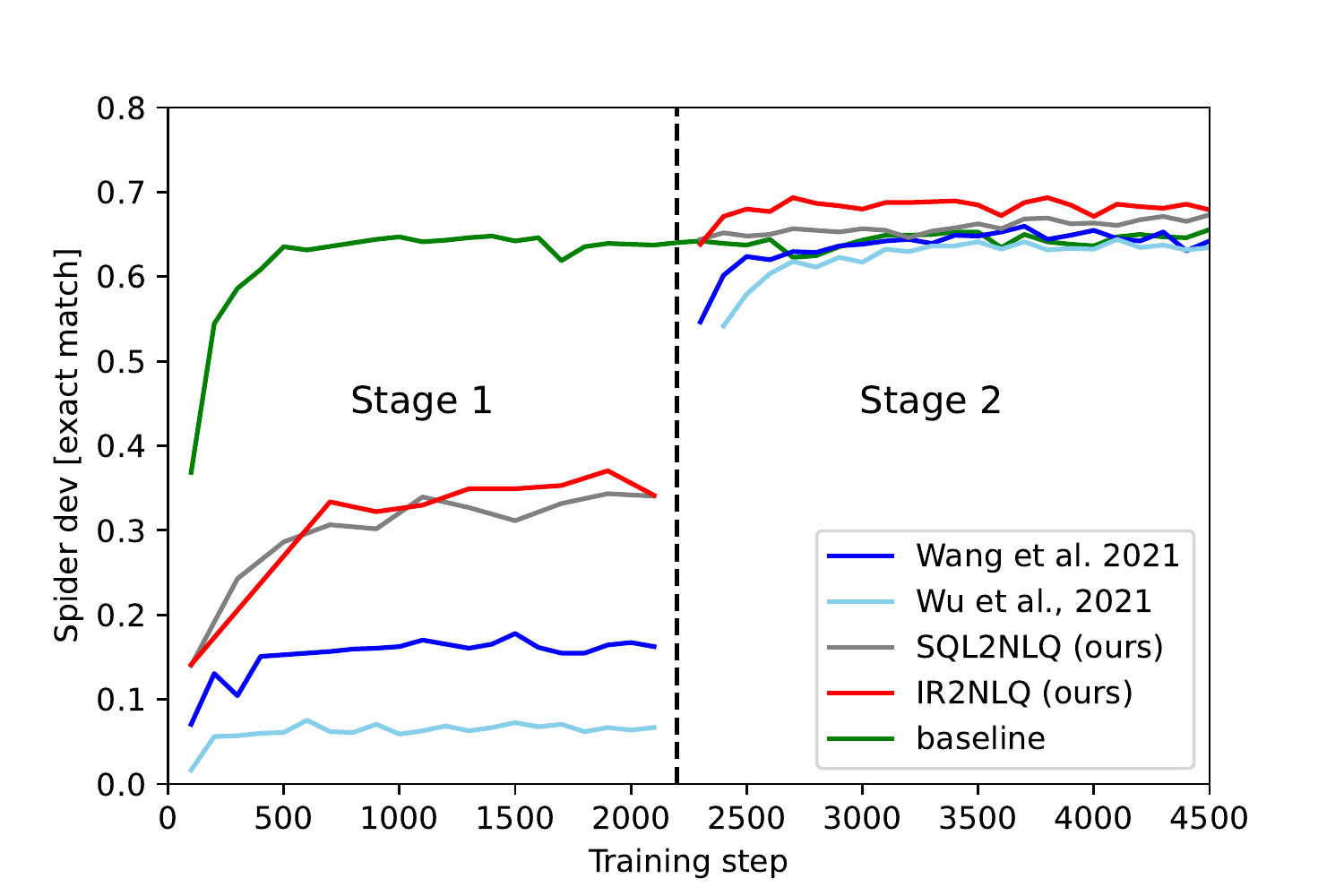}
     \caption{Training dynamics comparison of T5-Large with different synthetic data. The baseline model uses Spider real data only.  IR2NLQ and SQL2NLQ are our synthetic data with and without IR during NLQ generation.  We compare with previous synthetic datasets \cite{wu-etal-2021-data, wang-etal-2021-learning-synthesize}. We use synthetic data for stage-1 training and real data for stage-2 training.
    }

        \label{fig:assessment_ours}
\vspacereduce
\end{figure}

As a pilot study, we use T5-Large PICARD as the baseline parser to examine the synthetic data quality. As shown in Figure \ref{fig:assessment_ours}, the exact match (EM) accuracy on both synthetic datasets are less than 0.2 during Stage 1, in contrast to 0.6 with Spider training data only. This gap indicates the limited transferability from existing synthetic data to real data. Further finetuning on Spider training data in Stage 2 does not improve the baseline model. However, our synthetic data (IR2NLQ and SQL2NLQ) show better performance on these two stages. In the next sections, we reveal the synthetic data problems and detail our proposed method.




\subsection{Synthetic Data Quality Analysis}
We analyzed the previous synthesis methodologies and identified a few probable causes for obsolescence. 

\subsubsection{Illogical Synthetic SQLs from Invalid Grammars or Templates.}
Both \citet{wang-etal-2021-learning-synthesize} and \citet{wu-etal-2021-data} adopted context-free grammars to generate SQL queries. The CFG designed by \citet{wu-etal-2021-data} is constrained and they limited SQL generation to one table. While \citet{wang-etal-2021-learning-synthesize} designed flexible grammars, they neglected the constraints between operators and column types. This neglect leads to mistakes such as \texttt{SUM(student.name)}, where an aggregation operator is applied to a text column. 

Furthermore, PCFG generated SQL queries often failed to capture foreign-key and key relations between columns. This leads to invalid SQLs such as \texttt{SELECT name, age FROM student INTERSECT SELECT address FROM teacher}, where it intersects two sub-queries with different number of columns. In fact, designing a grammar to produce high coverage and logical SQLs is a difficult task due to the implicit dependencies of SQL elements. 

Alternatively, SQL templates extracted from training data better preserves column typing information \cite{zhong-etal-2020-grounded}. This approach drastically reduces the invalid SQLs caused by a misalignment between operators and column types. However, existing work still misses the critical key relations in the templates.




\subsubsection{Over-Complex SQLs from Arbitrary Multi-table Joins.} 
When SQLs are materialized, the column/table selection from existing work is independent and result in SQL queries with unnecessary complexity. 
Those queries often have unclear intent and thus are difficult to be correctly translated to natural language questions.
For instance, a simple template in Table \ref{fig:sql-length-example} that requires only two columns can be turned into a complicated and nonsensical SQL query with three table joins.


\subsubsection{Language Gap between SQL and NLQ.}
Recent work typically trains a sequence-to-sequence model to obtain corresponding natural language queries (NLQ) from synthetic SQLs \cite{wang-etal-2021-learning-synthesize,shi2021learning}. 
The gap between SQL-NLQ pairs are well recognized in text-to-SQL task and intermediate representation (IR) is commonly used to reduce such mismatch \cite{DBLP:journals/corr/abs-2109-05153,DBLP:conf/acl/GuoZGXLLZ19,DBLP:conf/emnlp/YuYYZWLR18,shi2021learning}.
However, the reverse of the source and target in SQL-to-text brings in its own challenge, such as incorrect references for \texttt{SELECT *}, missing conditions within long and complex SQL queries, and misinterpretation of ORDER phrases. 


\begin{figure}
\centering
\includegraphics[width=0.48\textwidth]
{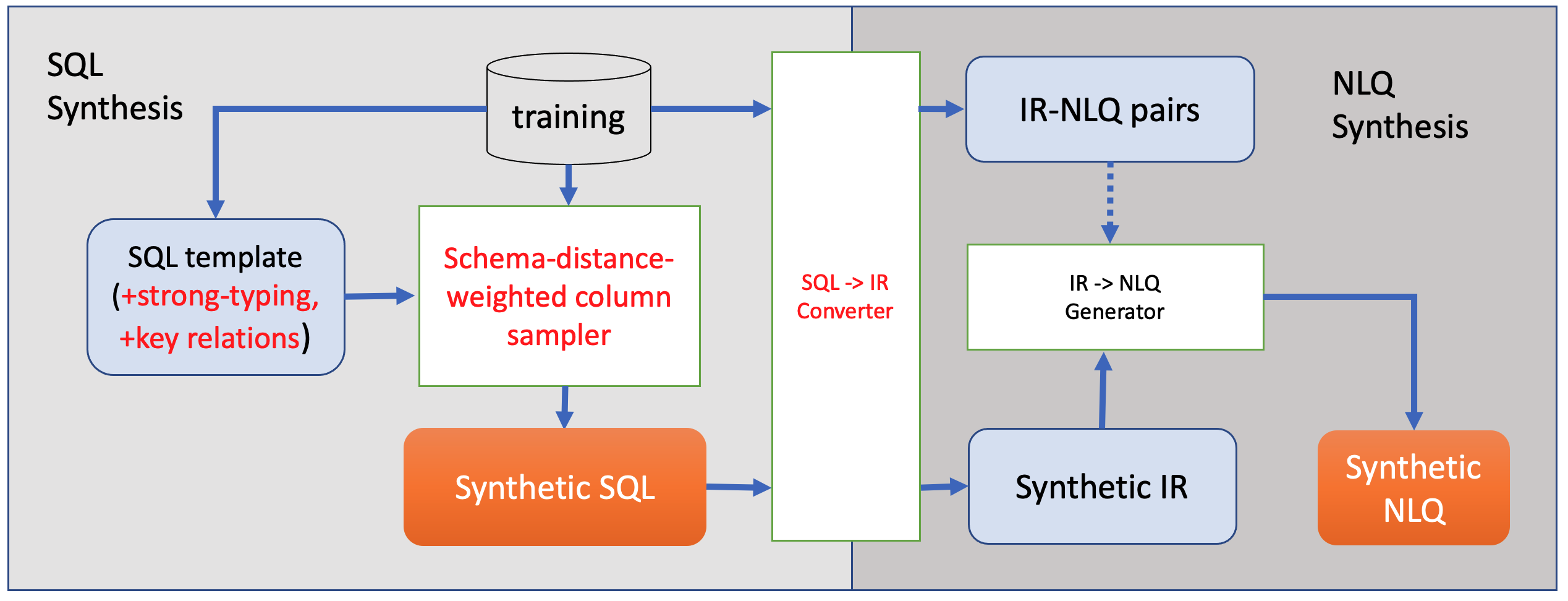}
\caption{Our NLQ-SQL synthesis framework. Novel components include strong-typing, key relations, schema-distance-weighted column sampler, and SQL~$\rightarrow$~IR converter.} 
\label{fig:synthesis-framework}
\vspacereduce
\end{figure}

\vspacesection
\section{Proposed Method}

This section outlines our proposed synthesis pipeline (Figure~\ref{fig:synthesis-framework}).
We follow the template based SQL synthesis approach similar to  \cite{zhong-etal-2020-grounded, zhang-etal-2019-editing} 
and generate corresponding NLQ with a sequence-to-sequence model. 
We address the generation problems reviewed in the previous sections by
\begin{itemize}
\item Introducing strong typing and encoding the key relation with the extracted templates for more logical SQLs.
\item Proposing a schema distance weighted column sampling strategy to avoid over-complex joins.
\item An improved IR to bridge the gap between SQL and natural language questions specifically for SQL-to-text.
\end{itemize}


\begin{table}[h!]
	\centering
	\caption{Our modifications for template extraction: strong typing is highlighted in blue and key relation preservation is highlighted in pink.}
	\label{fig:sql-example}
	\resizebox{0.95\columnwidth}{!}{
	\begin{tabular}{ll}
		\toprule
		\multirow{2}{*}{\textbf{SQL}} & \texttt{SELECT artist\_name FROM song INTERSECT}\\
																 & \texttt{SELECT artist\_name FROM artist}\\
		\midrule
		\textbf{Previous} & \texttt{SELECT col1\_key INTERSECT col2\_key} \\
		\textbf{Ours} & \texttt{SELECT col1\_\hlc[Cyan!50!]{textkey} INTERSECT col2\_\hlc[Cyan!50]{textkey}\_\hlc[Magenta!50]{fk1}} \\
		\bottomrule
	\end{tabular}
	}
\end{table}

\subsection{SQL Synthesis}
To create new SQLs on training data schemas, we utilize a template-based approach 
following \citet{zhong-etal-2020-grounded}:
First, a pool of SQL templates are created by normalizing the schema-related mentions (column and value) and removing JOIN phrases.
During SQL generation, a template is sampled based on the training distribution and columns are sampled with constraints to fill in the normalized slots of the template. We highlight several improvement made to the existing approach.


\subsubsection{Strong typing.} 
When normalizing columns, we enforce strong typing of a template by enriching and preserving the data type (e.g., text, number, date, etc) as well as key identity (key or not) for each column. For example, in Table \ref{fig:sql-example}, we use \texttt{textkey} instead of \texttt{key} to normalize \texttt{artist\_name} because operators such as \texttt{MAX} can be applied to number key but usually not to other text key. 



\subsubsection{Template Key Relationship Preservation.}
A foreign key is a column in a table referring to the primary key (unique identifier) of another table. 
In multiple table join scenarios, key and foreign key are the most common columns to be joined on.
Restricting a column to be a foreign key to another key column is critical for a SQL to be valid especially in the following two cases: 
1) queries including \texttt{INTERSECT, EXCEPT, UNION} and 
2) queries that contains nested queries in \texttt{WHERE} conditions.  
For instance, the query in Table \ref{fig:sql-example} implied the constraint that \texttt{song.artist\_name} should be a subset of \texttt{artist.artist\_name}.
 \texttt{FK1} in the template captures the constraint of key relationship between the two \texttt{artist\_name} columns, which prevents the template from generating nonsensical queries such as \texttt{SELECT \textbf{gender} FROM artist INTERSECT SELECT \textbf{country} FROM artist}. 

\subsubsection{Schema-distance-weighted Column Sampling.} 

To mitigate the issue of arbitrary multi-table joins, we implement a weighted sampling function biased toward columns that are close, in terms of table distance, to the columns already selected in a SQL template.

For a given database $d$, we first establish an undirected graph for all the tables in $d$. Each table represents a node in the graph. The distance between any two tables, $e(\cdot,\cdot)$, is the least number of joins necessary to join the two tables (i.e. shortest path distance) under the restriction that table join can only take place with qualified primary key and foreign key pairs. See Appendix A for more details.



Define a template $t$ as $(q, \mathbf c, \mathbf v)$ where $q$ is the flat template string, $\mathbf c = [c_1,\dotsc, c_m] $ is the set of column placeholders and $\mathbf v = [v_1,\dotsc, v_n] $ is the set of value placeholders in $q$. Denote $T_c$ to represent the table that contains column $c$ and $S_d(\tau)$ as the set of columns in $d$ with the \textit{strong type} $\tau$. Given a template $t$ and a qualified database $d$, the fundamental algorithm of SQL synthesis is described in Algorithm \ref{alg:sql_syn}.

\begin{algorithm}[!t]

\algsetup{linenosize=\tiny}
\footnotesize 

\caption{Single SQL Synthesis with Schema-Weighted Column Sampling}
\label{alg:sql_syn}
\textbf{Input}: template $t=(q, \mathbf c, \mathbf v)$, database $d$, decay rate $\gamma$\\
\textbf{Output}: SQL query $y$
\begin{algorithmic}[1] 
    \STATE Let $y=q$
    \STATE Random sample $z_1$ from $S_d(\tau_{c_1})$ and replace $c_1$ with $z_1$ in $y$ \\
    \STATE Compute sampling weights $$w(z) = \begin{cases} 1 , & \text{if } T_z=T_{c_1}\\ \frac{1}{\gamma^{\delta_{c_1}(z)}}, & \text{o.w.} \end{cases}, \quad \forall z$$ where $\delta_{c}(z) = e(T_{c}, T_z)$\\
    \FOR{$c\leftarrow c_2:c_m$}
        \STATE Compute sampling distribution $$p(z) = \begin{cases}\frac{w(z)}{\sum \limits_{z': \tau_{z'} = \tau_c} w(z')}, & \text{if } \tau_z = \tau_c \\ 0, & \text{o.w.} \end{cases}$$
        \STATE Sample $z$ from $S_d(\tau_c)$ with $p$\\
        \STATE Replace $c$ with $z$ in $y$ \\
        \STATE Update sampling weights
        $$w(z) \leftarrow w(z) + \begin{cases} 1 , & \text{if } T_z=T_{c}\\ \frac{1}{\gamma^{\delta_{c}(z)}}, & \text{o.w.} \end{cases}, \quad \forall z$$
    \ENDFOR
    \FOR{$v\leftarrow v_1:v_n$}
        \STATE Identify relevant columns w.r.t. $v$ and retrieve a set of possible values for $v$ from the $d$\\
        \STATE Random sample one value from the set and replace $v$ with the value in $y$
    \ENDFOR
\end{algorithmic}
\end{algorithm}

\begin{table}[!b]
	\centering
	\caption{Random sampling vs our schema-distance-weighted column sampling for a given template. The former produced a query with three joins while ours have both columns from the same table. }
	\label{fig:sql-length-example}
	\resizebox{0.95\columnwidth}{!}{
	\begin{tabular}{ll}
		\toprule
		\textbf{Template} & \texttt{SELECT col1\_numberkey WHERE col2\_name = VALUE}\\
		\midrule
	\multirow{4}{*}{\textbf{Random}} & \texttt{SELECT T1.Club\_ID FROM \hlc[Grey!50]{club} AS T1 JOIN \hlc[Grey!50]{coach} as T2}\\
																	 & \texttt{ON T1.Club\_ID = T2.Club\_ID JOIN \hlc[Grey!50]{player\_coach} AS T3} \\
																	 & \texttt{ON T2.Coach\_ID = T3.Coach\_ID JOIN \hlc[Grey!50]{player} AS T4}\\
																	 & \texttt{on T3.Player\_ID = T4.Player\_ID where T4.Rank = "3rd"} \\[5pt]
		\textbf{Ours} & \texttt{SELECT Club\_ID FROM \hlc[Grey!50]{club} WHERE Club\_Name="AIK"} \\
		\bottomrule
	\end{tabular}
	}
	\vspacereduce
\end{table}

The intuition behind the schema-weighted column sampling algorithm is as follows: after we select the first column for the given template, we want to choose other columns in the database that are more relevant to the first column, so as to boost the chance of synthesizing more realistic SQL queries. We do so by sampling columns, for the remaining column placeholders in the template, according to a particular sampling probability, which is a monotonically decreasing function of the edge value in the table graph for type-qualified \textit{column candidates}, and 0 for non-qualified \textit{column candidate}. Such implementation is motivated from the observation that over-lengthy SQLs resulted from multiple tables joins are rare in real world scenarios under the only-join-on-primary-key-foreign-key assumption. Table \ref{fig:sql-length-example} shows an example of how adopting the schema-weighted sampling can help reduce the unrealistic SQLs in the random case.

\begin{figure}[!h]
\vspace{-1em}
\centering
\includegraphics[height=0.25\textwidth]
{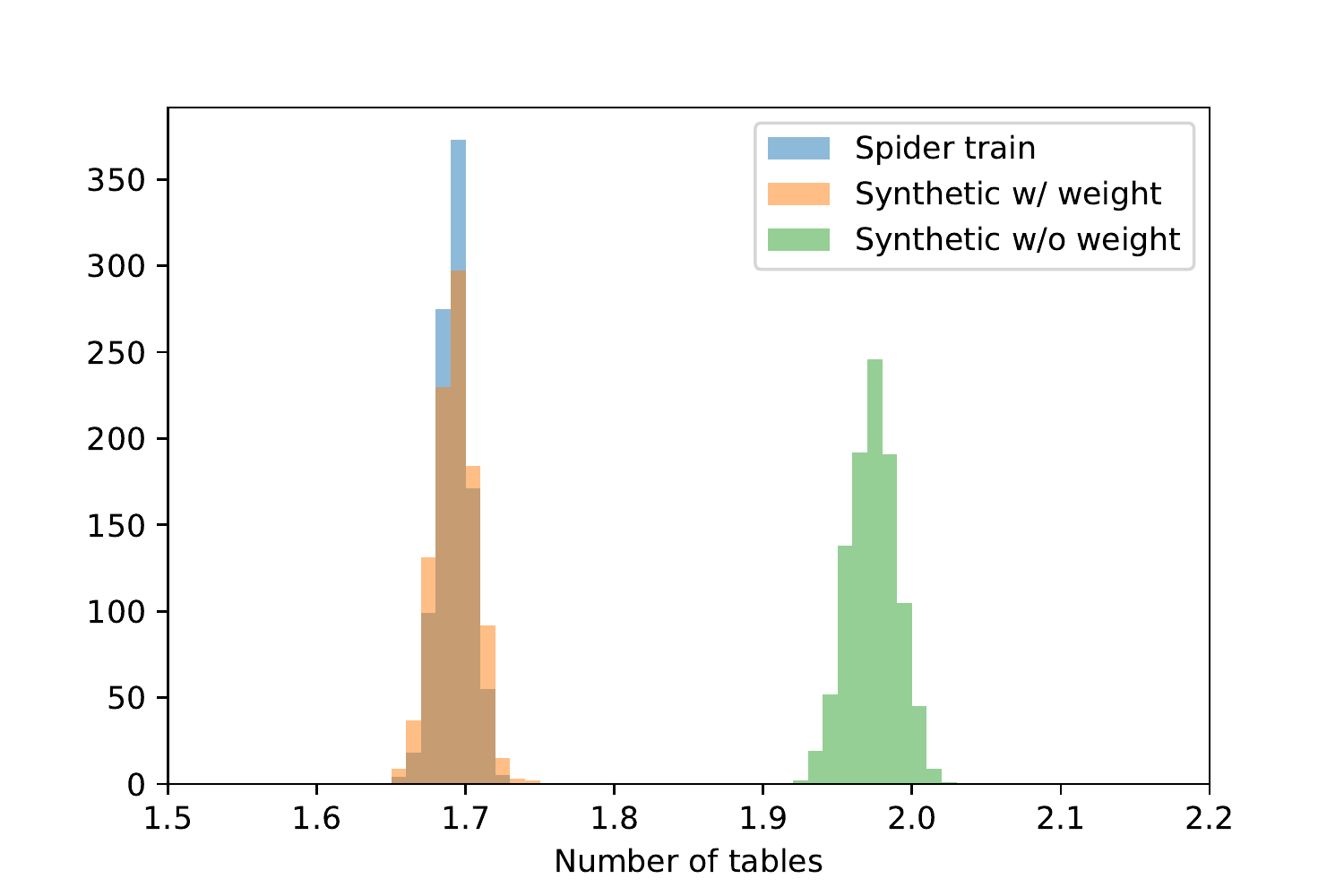}
\caption{
Histogram of the average table
count (i.e. umber of joins) for three types of datasets with $\gamma=5$. 
Our schema-distance-weighted column sampling reduces the table number of synthetic SQLs and better matches the training distribution.
} 
\label{fig:table-length}
\vspacereduce
\end{figure}

In Algorithm \ref{alg:sql_syn}, the input $\gamma$ is the hyperparameter that controls the decay rate in the sampling probability for qualified columns. By selecting an appropriate value for $\gamma$ ($\gamma=5$), the average table count in our synthetic data constructed from the schema-weighted column sampling method is close to that in the real Spider benchmark as shown in Figure \ref{fig:table-length}, while the random column sampling mechanism tend to generate SQLs that are overly complicated. See Appendix A for the experiment details.





\begin{table}[!t]
	\centering
	\caption{IR examples that illustrate the examples of removing tables, enriching \texttt{*} columns, specifying most/least intent, removing redundant \texttt{GROUP BY}. Unwanted intents are in grey, redundant intents are in green. Texts related to IR operations are highlighted with yellow.}
	\label{fig:ir1}
	\resizebox{\columnwidth}{!}{
	\begin{tabular}{lll}
		\toprule
		\multirow{4}{*}{\textbf{EX1}} & \textbf{SQL} & \texttt{SELECT T1.name {\color{gray}FROM student AS T1 JOIN} \hlc[yellow]{has\_pet} {\color{gray}AS T2}}\\
																	& & \texttt{{\color{gray}\;\;ON T1.student\_id = T2.has\_pet.student\_id}} \\ [2pt]
																	& \textbf{IR}  & \texttt{SELECT name of student FROM \hlc[yellow]{has\_pet}} \\ [2pt]
																	& \textbf{NLQ} & \Large{Find the name of students who \hlc[yellow]{have pets}.} \\ [2pt]
		\midrule
		\multirow{5}{*}{\textbf{EX2}} & \textbf{SQL} & \texttt{SELECT T2.name, \hlc[yellow]{count(*)} {\color{gray}FROM concert AS T1 JOIN stadium AS}}\\
																	& & \texttt{{\color{gray}\;\;T2 ON T1.stadium\_id = T2.stadium\_id} GROUP BY T1.stadium\_id} \\ [2pt]
																	& \textbf{IR}  & \texttt{SELECT name of stadium, \hlc[yellow]{Count ( record of concert )}} \\
																	& & \texttt{\;\;GROUP BY ( stadium\_id of concert )} \\ [2pt]
																	& \textbf{NLQ} & \Large{Show the stadium name and \hlc[yellow]{the number of concerts} in each stadium.} \\ [2pt]
		\midrule
		\multirow{8}{*}{\textbf{EX3}} & \textbf{SQL} & \texttt{SELECT T1.neighbourhood\_name {\color{gray}neighbourhood AS T1 JOIN}} \\
																	& & \texttt{{\color{gray}\;\;business AS T2 ON T1.business\_id = T2.business\_id}} \\
																	& & \texttt{\;\;WHERE T2.city = "Madison" \hlc[green]{GROUP BY T1.neighbourhood\_name}}\\
																	& & \texttt{\;\;\hlc[yellow]{ORDER BY COUNT ( DISTINCE T2.name ) DESC LIMIT 1}} \\ [2pt]
																	& \textbf{IR}  & \texttt{SELECT neighbourhood\_name of neighbourhood} \\
																	& & \texttt{\;\;\hlc[yellow]{WITH most Count ( DISTINCT name of business )}} \\
																	& & \texttt{\;\;WHERE city of business = "Madison"}\\ [2pt]
																	& \textbf{NLQ} & \Large{Which neighbourhood \hlc[yellow]{has the most number of businesses} in Madison?} \\ [2pt]
		\midrule
		\multirow{4}{*}{\textbf{EX6}} & \textbf{SQL} & \texttt{SELECT T2.name {\color{gray}FROM USER AS T2 JOIN review AS T1}} \\
																	& & \texttt{\;\;{\color{gray}ON T2.user\_id = T1.user\_id} \hlc[green]{GROUP BY T2.name}} \\
																	& & \texttt{\;\;HAVING AVG (T1.rating) < 3} \\ [2pt]
																	& \textbf{IR}  & \texttt{SELECT \hlc[yellow]{EACH ( name of user )} WITH Avg} \\
																	& & \texttt{\;\;( rating of review ) < 3} \\ [2pt]
																	& \textbf{NLQ} & \Large{Find \hlc[yellow]{users} whose average review rating is below 3.} \\ [2pt]
		\bottomrule
	\end{tabular}
	}
\end{table}


\subsection{NLQ Synthesis}
Intermediate representation (IR) has been employed to simplify the SQL query with minimum information loss \cite{gan-etal-2021-natural-sql,guo-etal-2019-towards,DBLP:journals/corr/abs-2109-05153,DBLP:conf/acl/GuoZGXLLZ19,DBLP:conf/emnlp/YuYYZWLR18,shi2021learning}. Common operations include removing \texttt{FROM/JOIN} clauses and \texttt{GROUP BY} clauses, and merging \texttt{WHERE} clauses and \texttt{HAVING} clauses. Previous works find the use of IR often improves text-to-SQL performance. 

In this section, we explore whether the SQL-to-text generation could also benefit from an IR. According to a prior research by \citet{wu-etal-2021-data},  altering the query's linearization order could already affect the synthetic text quality. The objective of an IR here is to convert SQL to a representation that more closely resembles the NLQ. This 
conversion involves both simplifications (such as removal of redundant information) and specification (such as introducing information using heuristics).

We outline the main new rules to transform SQLs into IRs and explain the rationale (examples in Table \ref{fig:ir1}):

\begin{itemize}
  \setlength\itemsep{0em}

    \item  Only drop tables in the \texttt{FROM/JOIN} phrase if they appear in other SQL elements (\textbf{EX2-EX4}). Removal of tables can simplify queries but tables in \texttt{JOIN} can also behave as filters and need to be preserved to avoid information loss (\textbf{EX1}).
    \item Replace \texttt{*} in \texttt{count(*)} with the table whose columns in \texttt{JOIN} act as foreign key 
    to provide explicit context for counting. 
    This is because, in multi-table join queries, foreign key represents the \texttt{many} of the one-to-many relations and thus the rows from the table is more meaningful to be aggregated (see \textbf{EX2} replaces\texttt{*}with \texttt{concert} rather than \texttt{stadium}). 
    \item When SQL contains \texttt{ORDER BY COUNT (...) LIMIT ...} , rewrite the query to explicitly express the most or least intent for better intent alignment (\textbf{EX3}).
    \item Drop \texttt{GROUP BY} phrase if the column grouped by appears in \texttt{SELECT} and attach \texttt{EACH} to the specific column if the query does not express the most/least intent (see \texttt{GROUP} dropped in \textbf{EX3 - EX4} but not \textbf{EX2}). This aims to distinguish SQLs with \texttt{GROUP BY} and \texttt{SELECT} on the same column from those without \texttt{SELECT}.
\end{itemize}
Similar to previous IR designs, we also removed repeated text in \texttt{EXCEPT/INTERSECT/UNION} queries and made lexical adjustments. 


\vspacesection
\section{Experiments}
We conduct experiment on the challenging Spider benchmark \cite{yu-etal-2018-spider}, which contains various complex SQL statements and realistic cross-database evaluation setting. We demonstrate the effectiveness of our data synthesis framework from both text-to-SQL and SQL-to-text. 



\subsubsection{Spider Benchmark} Spider \cite{yu-etal-2018-spider} is a large-scale text-to-SQL dataset, it has 10,181 annotated questions, 5693 unique complex SQLs and 200 databases with multiple tables. It also contains datasets from previous works, such as Restaurants \cite{tang-mooney-2000-automated, 10.1145/604045.604070}, GeoQuery \cite{10.5555/1864519.1864543}, Scholar \cite{iyer-etal-2017-learning}, Academic \cite{10.14778/2735461.2735468}, Yelp and IMDB \cite{https://doi.org/10.48550/arxiv.1702.01168}, which are compiled as \textbf{train-others}. 
The \textbf{train}/\textbf{train-others}/\textbf{dev}/\textbf{test} sets contain 7000/1659/1034/2147 examples and  
140/6/20/40 databases, respectively. 
Spider has a challenging and realistic evaluation setting, where SQL queries and databases do not appear across different splits, posing a generalization challenge for text-to-SQL semantic parser. Since Spider test set is not publicly available, we use dev set for evaluation and train-others for checkpoint selection. 

\subsubsection{Text-to-SQL Parser} We use T5-3B \cite{JMLR:v21:20-074} as our base parser, since previous work \cite{shaw-etal-2021-compositional} has shown that T5-3B can achieve competitive performance for Text-to-SQL semantic parsing. Recently, PICARD \cite{Scholak2021:PICARD} demonstrates that constraint decoding on top of T5-3B can produce state-of-the-art performance on Spider.

\subsubsection{SQL-to-Text and IR-to-Text Generator} We finetune a T5-large model on Spider training set for both SQL-to-text generation and IR-to-text generation, the best checkpoint is selected with the highest BLEU score on \textbf{train-others}. 


\subsubsection{Configurations} We adopt a two-stage text-to-SQL training mechanism \cite{wang-etal-2021-learning-synthesize} in our experiment. In the first stage, we use synthetic data only for model pre-finetuning. In the second stage, we initialize the model weights with the first stage checkpoint, and then finetune it on the real data only. Both stages share the same hyperparameters, we train T5 with Adafactor and learning rate of 1e-4, and use gradient accumulation batch size 2050 and 64 for T5-3B and T5-Large, respectively. Our experiments are based on NVIDIA A100-SXM4-40GB GPUs, we use beam size 5 and top-2 predictions for PICARD decoding.

\begin{table}[ht]
	\small
	\centering
    \caption{Comparison of the top-performing text-to-SQL models in Spider leaderboard, as well as models trained with synthetic data (where synthetic are generated by training schema only). We report exact set match (EM) and execution accuracy (EX) for Spider dev set. $\dagger$ means T5-3B is trained with database content. When trained with our synthetic data, 
    T5-3B model has 4.4 points of EM improvement; and T5-3B$^\dagger$ PICARD has 2.1 points of EX improvement.
    } 
    \resizebox{0.97\columnwidth}{!}{
    \begin{tabular}{lcc}
    \toprule
    \textbf{Model} & \textbf{EM} & \textbf{EX} \\
    \midrule
     DT-Fixup SQL-SP \cite{xu-etal-2021-optimizing} & 75.0 & - \\
     LGESQL + ELECTRA \cite{cao-etal-2021-lgesql} & 75.1 & - \\
     S2SQL + ELECTRA \cite{hui-etal-2022-s2sql} & \underline{76.4} & - \\
     DT-Fixup + Syn \cite{yang2021hierarchical} & \underline{76.4} & - \\
     T5-3B \cite{shaw-etal-2021-compositional} & 70.0 & - \\
     T5-3B + Syn data \cite{wu-etal-2021-data} & 69.1 & - \\
     T5-3B + Syn data \cite{wang-etal-2021-learning-synthesize} & 70.3 & - \\
     T5-3B + Syn data (ours) & 74.4 & - \\
     T5-3B + PICARD (Scholak {\etal}, 2021) & 74.1 & - \\
     T5-3B + PICARD + Syn data (ours) & \textbf{76.9} & - \\
     \midrule
     SmBoP + GraPPa \cite{rubin-berant-2021-smbop} & 69.5 & 71.1 \\
     GAP + NatSQL \cite{gan-etal-2021-natural-sql} & 73.7 & 75.0 \\
     T5-3B$^\dagger$ (Scholak {\etal}, 2021) & 71.5 & 74.4 \\
     T5-3B$^\dagger$ + Syn data (ours) & {74.5} & {78.6} \\
     T5-3B$^\dagger$ + PICARD (Scholak {\etal}, 2021) & \underline{75.5} & 79.3 \\
     RASAT + PICARD \cite{rasat} & 75.3 & \underline{80.5} \\
     T5-3B$^\dagger$ + PICARD + Syn data (ours) & \textbf{76.1} & \textbf{81.4} \\
     \bottomrule
    \end{tabular}}
    \label{tab:Spider_SOTA}
\vspacereduce
\end{table}

\subsection{Spider Results and Analysis}
The overall results\footnote{Some models do not predict cell values or access to database content, we leave `-' for EX.} are shown in Table \ref{tab:Spider_SOTA}, where we can see our synthetic data can further improve the state-of-the-art model and achieve the best results on Spider development set\footnote{Since the official test set is hidden, we have not received their evaluation results as of submission time}, including both exact set match and execution accuracy. 
Specifically, we have 4.4 points of EM score improvement on top of T5-3B model, while previous works \cite{wu-etal-2021-data, wang-etal-2021-learning-synthesize} have marginal gain or even hurt the performance, demonstrating the effectiveness of our proposed method. More importantly, T5-3B was proved to show SOTA or near SOTA performance on 21 knowledge grounding tasks \cite{UnifiedSKG}, our success of improving T5-3B with synthetic data for text-to-SQL can potentially generalize to other semantic parsing tasks with different logical forms. 

PICARD is an incremental parsing method for constraint decoding, which can reduce the syntax errors of language models for SQL generation. From Table \ref{tab:Spider_SOTA}, we see that T5-3B combined with PICARD and our synthetic data performs the best, implying the orthogonality of synthetic data augmentation and constraint coding. However, the gain of PICARD is reduced if we pre-finetune T5-3B with our synthetic data, for example, PICARD can improve T5-3B$^\dagger$ by 4 points of EM score, but only 1.6 points on top of our synthetic data. 

\begin{table}[t]
\small
\centering
    \caption{Generated NLQ quality evaluations on the Spider dev set between SQL$\rightarrow$ NLQ and SQL$\rightarrow$ IR$\rightarrow$ NLQ.  The BLEU \cite{papineni-etal-2002-bleu}, ROUGE \cite{lin-2004-rouge}, and BERT \cite{Zhang*2020BERTScore:} scores show that IR helps generate NLQs that are closer to the groundtruth.}
    \resizebox{0.95\columnwidth}{!}{
    \begin{tabular}{lccccc}
    \toprule
    \textbf{Settings} & \textbf{BLEU}  & \textbf{R-1} & \textbf{R-2} & \textbf{P-BERT} & $\textbf{R-BERT}$\\
    \midrule
    SQL$\rightarrow$ NLQ & 27.7 & 59.6 & 35.3 & 93.6 & 93.2\\
    SQL$\rightarrow$ IR$\rightarrow$ NLQ & 29.3 & 60.5 & 36.8 & 93.9 & 93.3 \\
     \bottomrule
    \end{tabular}}
    \label{tab:NLQ_generation}
\end{table}

In order to understand the effectiveness of our proposed method for both SQL and IR synthesis, we plot T5-Large training curves with different synthetic datasets in Figure \ref{fig:assessment_ours}. Compared with previous works \cite{wu-etal-2021-data, wang-etal-2021-learning-synthesize}, our synthetic data demonstrates significant improvement in stage-1, either from SQL$\rightarrow$NLQ or SQL$\rightarrow$IR$\rightarrow$NLQ, proving the high-quality of our synthesized SQLs. Additionally, with the help of IR, we can further boost the stage-2 performance. We also compare the generated NLQs with different automatic measurements in Table \ref{tab:NLQ_generation}, where we can see IR benefits the NLQ generation process and produces the text closer to groundtruth NLQs.

\subsection{Synthetic Data Efficiency}
In this section, we study the efficiency of our synthetic data framework from different aspects.


\begin{table}[t]
\small
\centering
    \caption{Text-to-SQL experiment with the few-shot setting, where we sampled a subset from the original Spider training set with size varying from 128 to 1024, then created synthetic data with templates only from the subset. \textbf{\# tmpl} and \textbf{\# syn} represent the number of templates and synthesized NLQ-SQL pairs for the corresponding training subset. We report exact set match on the Spider dev set.}
    \resizebox{0.95\columnwidth}{!}{
    \begin{tabular}{lcccccc}
    \toprule
     \multirow{3}{*}{\textbf{Model }} & \textit{ f}\textbf{-shot:} & \textbf{128}  & \textbf{256} & \textbf{512} & \textbf{1024} & \textbf{full (7k)}\\
     & \textbf{\# tmpl} & \textbf{68} & \textbf{116} & \textbf{205} & \textbf{318} & \textbf{746} \\
     & \textbf{\# syn} & \textbf{7839} & \textbf{10775} & \textbf{14457} & \textbf{17002} & \textbf{21851}\\
    \midrule
    \multirow{2}{*}{T5-3B} & real only & 19.1 & 32.3 & 43.6 & 53.2 & 70.0 \\
     & real + syn & 46.3 & 54.4 & 59.9 & 62.2 & 74.4 \\
     \bottomrule
    \end{tabular}}
    \label{tab:Spider_few_shot}
\end{table}

\subsubsection{Few-shot setting: How much real data do we need to rely on before achieving acceptable performance?} Since annotating text-to-SQL dataset takes extremely high human effort, in practice, it's hard to create a large-scale corpus with a limited annotation budget. Table \ref{tab:Spider_few_shot} presents the text-to-SQL semantic parsing results with a limited number of training examples, we also generate our synthetic data on top of the corresponding subset. Interestingly, as training size decrease from 7K to 128, our synthetic data becomes more essential, the performance gain increases from 4.4 points to 27.2 points. Even with only 512 training examples, our synthetic data can assist the T5-3B model to achieve $\sim$60\% accuracy level. These few-shot setting results are encouraging, as we can annotate a small-scale training set but still achieve acceptable performance with the help of synthetic data. 
\begin{figure}[h]
\vspace{-1em}
     \centering
     \includegraphics[height=0.25\textwidth]{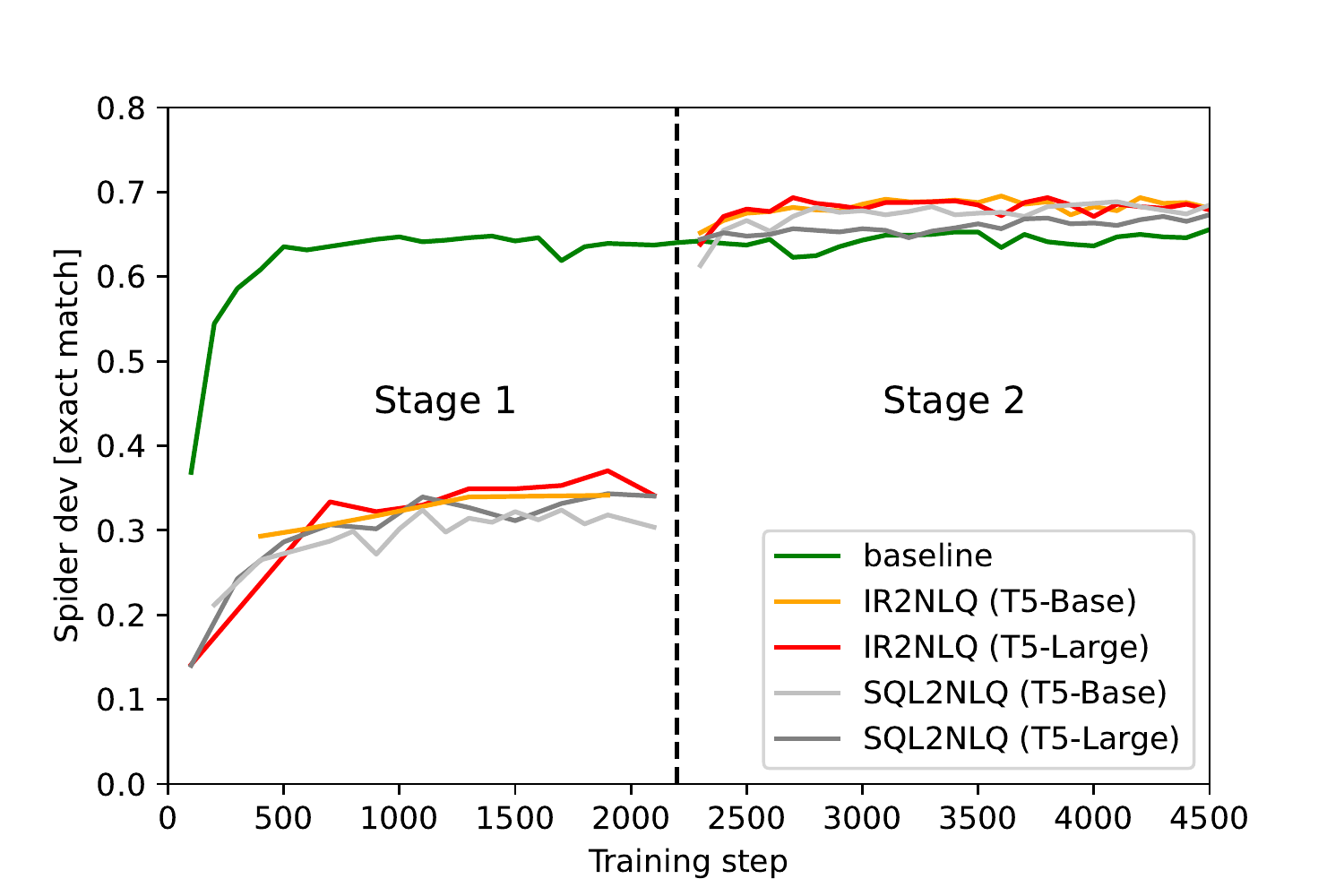}
     
        \caption{Comparison of different T5 model sizes for NLQ generations. On top of T5-Base (220M parameters) and T5-Large (770M parameters), we finetune generators for both SQL-to-text and IR-to-text, then evaluate the effectiveness with text-to-SQL semantic parsing in Spider. }
        \label{fig:assessment_generator}
\vspacereduce
\end{figure}

\subsubsection{Generator size: How big of the generator model do we need to use to produce high-quality NLQs?} Since our proposed IR is to reduce the gap between NLQ and SQL, we hypothesize that the NLQ generation process should have less reliance on model size. In our data synthesis framework, after generating the SQLs and the corresponding IRs, we use T5-Large by default for NLQ generations. However, our IR is designed to simplify the translation process from SQL to NLQ, we think this should be a relatively easy task. As shown in Figure \ref{fig:assessment_generator}, even with smaller T5-Base as generator, our synthetic data (with IR2NLQ) still presents comparable performance, implying the effectiveness and robustness of our proposed IR. As comparison, SQL2NLQ has larger divergence between T5-Large and T5-Base, indicating some difficulty of translating SQL to NLQ.

\subsubsection{Seen schema: How good of the synthetic data if we consider a broader coverage of database schema?} Since the cross-database evaluation setting presents generalization challenge for text-to-SQL parsers, our synthetic framework can potentially overcome this by utilizing more public database schemas, or even ones that can implicitly cover the evaluation set. In addition to using schema from training set, we can take advantage of more public schemas for data synthesis, for example, WikiTables \cite{Bhagavatula2015TabELEL}, GitTables \cite{hulsebos2021gittables}, WikiSQL \cite{zhongSeq2SQL2017} and SQL tutorial websites, some of them are even schema source for Spider benchmark. We simply added 20 databases from dev set into our synthetic data generation, then trained text-to-SQL parser on top of T5-Large. With this setting, we observed $\sim$2 points of performance improvement compared to that with training schema only. This pilot study implies the potential helpfulness of synthesizing data with targeting database schemas to further improve the downstream performance.

\subsubsection{Single-table: How effective is our method on the single-table text-to-SQL parsing?} Although our SQL synthesis is mainly designed for multi-table operations, it should also be compatible with the single table, but with foreign key preservation ineffective. WikiSQL \cite{zhongSeq2SQL2017} and SQUALL \cite{shi-etal-2020-potential} are two popular datasets for single-table text-to-SQL parsing. Compared to multi-table case, the single-table is much easier, for example, most text-to-SQL parsers are above 90\% accuracy level in WikiSQL\footnote{\url{https://github.com/salesforce/WikiSQL}}. We took a relatively challenging SQUALL dataset for experiment, from 9K training examples, we created 30K synthetic NLQ-SQL pairs. With the original training data, T5-Base can achieve 69.2\% execution accuracy, after augmenting with our synthetic data, the accuracy is improved to 69.7\%. The performance gain is not significant, we hypothesize several reasons: 1) foreign key relationship is not applicable in single table, but critical to our data synthesis framework; 2) 9k examples are sufficient for model training, especially for SQLs without JOIN clause, therefore the effect of synthetic data is further diluted. 

\vspacesection
\section{Conclusion}
In this work, we proposed a data synthesis framework for text-to-SQL semantic parsing. After incorporating key relationships from schema, imposing strong typing, conducting schema-distance-weighted column sampling and bridging SQL~$\rightarrow$~NLQ generation with intermediate representation, we synthesized high-quality dataset that can further improve the state-of-the-art parser on Spider benchmark. We also revealed the efficiency of the synthetic data and pointed out the potential usefulness of reducing human annotations for text-to-SQL parsing.

\bibliography{aaai23}

\clearpage
\onecolumn
\subsection{Appendix A: Details on Schema-distance-weighted Column Sampling}

\subsubsection{Table Distance.}

For a given database $d$, we first establish an undirected graph for all the tables in $d$. We can then compute the distance between any two tables, $e(\cdot,\cdot)$, defined as the least number of joins necessary to join the two tables under the restriction that table join can only take place with qualified primary key and foreign key information. In other words, we disable arbitrary join of two tables if they lack key and foreign key relationship.

We give some examples using one of the databases (id: \texttt{college\_1}) in the Spider benchmark, as shown in Table \ref{tab:exdb}.
\begin{itemize}
	\item $e(\text{T1}, \text{T2})=1$ because the column \verb|class code| in table \textit{class} (T1) is a foreign key in table \textit{course} (T2). We can also observe from the table graph in Figure \ref{fig:graph}: there is a direct path between table node \textit{class} and table node \textit{course}.
	\item $e(\text{T2}, \text{T7})=2$ since we first need to join table \textit{course} (T2) with table \textit{department} (T3), followed by joining table \textit{department} with table \textit{student} (T7). Note that even though we can also join using the path $T2\rightarrow T1 \rightarrow T5 \rightarrow T7$, this is not the \textbf{\textit{least}} number of joins between the two tables.
\end{itemize}

\begin{table*}[!h]
	\setlength\extrarowheight{5pt}
	\centering
	\caption{Example database (id: \texttt{college\_1})}
	\label{tab:exdb}
	\begin{tabular}{ccccc}
		\multirow{2}{*}{Alias} & \multirow{2}{*}{Table Name} & \multirow{2}{*}{Primary Key} & \multicolumn{2}{c}{Foreign Key} \\ [5pt] \cline{4-5}
																 & & & Table & Column \\
		\toprule
		T1 & class & \verb|class code| & enroll & \verb|class code| \\
		\hline
		T2 & course & \verb|course code| & class & \verb|class code| \\
		\hline
		\multirow{3}{*}{T3} & \multirow{3}{*}{department} & \multirow{3}{*}{\texttt{department code}} & course  & \verb|department code| \\
												&&& professor & \verb|department code| \\
												&&& student & \verb|department code| \\
												\hline
		\multirow{3}{*}{T4} & \multirow{3}{*}{employee} & \multirow{3}{*}{\texttt{employee number}} & class  & \verb|professor employee number| \\
												&&& department & \verb|employee number| \\
												&&& professor & \verb|employee number| \\
												\hline
		T5 & enroll & - & - & - \\
		\hline
		T6 & professor & - & - & - \\
		\hline
		T7 & student & \verb|student num| & enroll & \verb|student number| \\
		\bottomrule
	\end{tabular}
\vspacereduce
\end{table*}

\begin{figure*}[!h]
	\centering

	\resizebox{0.69\textwidth}{!}{
		\begin{tikzpicture}[auto, thick,
			table/.style = {circle, inner sep=0pt, outer sep=0pt, minimum width=60pt, draw=orange!80},
			]
			\foreach \place/\name/\label in {{(0,0)/T2/course}, {(0,-4)/T1/class}, {(4,3)/T3/department}, {(4,-2)/T4/employee}, {(-4,-1.5)/T5/enroll}, {(8,0)/T6/professor}, {(-2,3)/T7/student}}
				\node[table] (\name) at \place {\begin{tabular}{c} \label \\ \name \end{tabular}};
			\foreach \source/\dest in {T1/T5, T2/T1, T3/T2, T3/T6, T3/T7, T4/T1, T4/T3, T4/T6, T7/T5}
				\path (\source) edge (\dest);
		\end{tikzpicture}
	}
	\caption{Example table graph (id: \texttt{college\_1})}
	\label{fig:graph}
\end{figure*}

The reason we introduce the concept of \textit{table distance} is that we want to leverage this value to promote table joins with appropriate relationships while discouraging illogical joins when two tables are irrelevant. During the process of column sampling, we will choose columns that have smaller table distance with the other columns that have already been selected with the objective to create more realistic synthetic SQL queries. In the example above, assume we have first sampled a column from the table \textit{student} (T7). For the next column placeholder, we are more likely to sample a column from table \textit{enroll} (T5) than table \textit{professor} (T6) --- it is more natural to ask questions like "how many students enrolled in class X" compared to asking "how many students enrolled in classes taught by professors who were employed before year YYYY".

\subsubsection{Value of $\gamma$ in Algorithm \ref{alg:sql_syn}.}

Recall that in Algorithm \ref{alg:sql_syn}, $\gamma$ is a hyperparameter that controls the decay rate in the sampling probability for columns that are farther away from the columns that have already been selected. Under the restricted join condition, we look at the number of tables in a query as a proxy to the table distance. To determine the value of $\gamma$, we randomly sample 7000 synthetic SQL queries with replacement and calculate the average number of tables from the samples. We repeat this process for 1000 times and plot the distribution. Then we perform the same steps for the real Spider training data. We chose $\gamma$ so that the distribution of the average number of tables in the synthetic data is close to the real data. This helps prevent generating over-simplified or over-complicated SQL queries.

Based on this experiment, we chose $\gamma$ to be 5 for the Spider benchmark. Figure \ref{fig:table-length} displays the distribution for three types of datasets: Spider training, synthetic dataset with schema-distance-weighted column sampling, and synthetic dataset with random column sampling. The figure demonstrates that the weighted sampling process, which provides an interface to tune the value of $\gamma$, can generate synthetic SQL queries that better match the real training data.

%

\end{document}